\documentclass[conference]{IEEEtran}
\IEEEoverridecommandlockouts

\usepackage{cite}
\usepackage{amsmath,amssymb,amsfonts}
\usepackage{graphicx}
\usepackage{textcomp}
\usepackage{subfig}
\usepackage{algorithm}
\usepackage{algorithmicx, program, algpseudocode}
\ifCLASSINFOpdf
\else
\fi
\begin{document}
%
\title{A Novel Application of Image-to-Image Translation: Chromosome Straightening Framework by Learning from a Single Image}

\author{\IEEEauthorblockN{Sifan Song\IEEEauthorrefmark{1},
Daiyun Huang\IEEEauthorrefmark{2},
Yalun Hu\IEEEauthorrefmark{3}, 
Chunxiao Yang\IEEEauthorrefmark{4},
Jia Meng\IEEEauthorrefmark{2},
Fei Ma\IEEEauthorrefmark{5},
Frans Coenen\IEEEauthorrefmark{6}, \\
Jiaming Zhang\IEEEauthorrefmark{7} and
Jionglong Su\IEEEauthorrefmark{8}\thanks{This work has been submitted to the IEEE for possible publication. Copyright may be transferred without notice, after which this version may no longer be accessible.}}
\IEEEauthorblockA{\IEEEauthorrefmark{1}Department of Mathematical Sciences, Xi'an Jiaotong-Liverpool University, Suzhou 215123, China}
\IEEEauthorblockA{\IEEEauthorrefmark{2}Department of Biological Sciences, Xi'an Jiaotong-Liverpool University, Suzhou 215123, China}
\IEEEauthorblockA{\IEEEauthorrefmark{3}Department of Applying Algorithm Engineering, Synsense Ltd. Chengdu 610041, China}
\IEEEauthorblockA{\IEEEauthorrefmark{4}Suzhou Sano Precision Medicine Ltd., Suzhou 215123, China}
\IEEEauthorblockA{\IEEEauthorrefmark{5}Department of Applied Mathematics, Xi'an Jiaotong-Liverpool University, Suzhou 215123, China}
\IEEEauthorblockA{\IEEEauthorrefmark{6}Department of Computer Science, University of Liverpool, Liverpool, L69 3BX, UK}
\IEEEauthorblockA{\IEEEauthorrefmark{7}Institute of Robotics and Intelligent Manufacturing, The Chinese University of Hong Kong (Shenzhen) \\and Shenzhen Institute of Artificial Intelligence and Robotics for Society, Shenzhen, China \\zhangjiaming@cuhk.edu.cn \{corresponding author\}}
\IEEEauthorblockA{\IEEEauthorrefmark{8}School of AI and Advanced Computing, XJTLU Entrepreneur College (Taicang), Xi'an Jiaotong-Liverpool University, \\Suzhou 215123, China, Jionglong.Su@xjtlu.edu.cn \{corresponding author\}}
}


\maketitle

\begin{abstract}
In medical imaging, chromosome straightening plays a significant role in the pathological study of chromosomes and in the development of cytogenetic maps. Whereas different approaches exist for the straightening task, typically geometric algorithms are used whose outputs are characterized by jagged edges or fragments with discontinued banding patterns. To address the flaws in the geometric algorithms, we propose a novel framework based on image-to-image translation to learn a pertinent mapping dependence for synthesizing straightened chromosomes with uninterrupted banding patterns and preserved details. In addition, to avoid the pitfall of deficient input chromosomes, we construct an augmented dataset using only one single curved chromosome image for training models. Based on this framework, we apply two popular image-to-image translation architectures, U-shape networks and conditional generative adversarial networks, to assess its efficacy. Experiments on a dataset comprised of 642 real-world chromosomes demonstrate the superiority of our framework, as compared to the geometric method in straightening performance, by rendering realistic and continued chromosome details. Furthermore, our straightened results improve the chromosome classification by 0.98\%-1.39\% mean accuracy.
\end{abstract}


%
\IEEEpeerreviewmaketitle

\section{Introduction}
\PARstart{T}{here} are 23 pairs of chromosomes in a normal human cell, comprised of 22 autosomes pairs (Type 1 to Type 22) and a pair of sex chromosomes (XX in females and XY in males). In the metaphase of cell division, the chromosomes become condensed and can be stained by the Giemsa banding technique \cite{speicher2005new} for observation under optical microscopes. The unique presence of light and dark regions (banding patterns) of different chromosome types are integrated into bars as cytogenetic maps. These banding patterns provide essential evidence for uncovering chromatin localization, genetic defects, and abnormal breakages \cite{saidzhafarova2009molecular}. For instance, human genetic diseases, such as cri-du-chat syndrome \cite{hills2006cri} and Pallister-Killian mosaic syndrome \cite{kostanecka2012developmental}, can be diagnosed by identifying structural abnormalities in chromosomes.

\begin{figure}[htbp]
\centering
\includegraphics[width=3.5in]{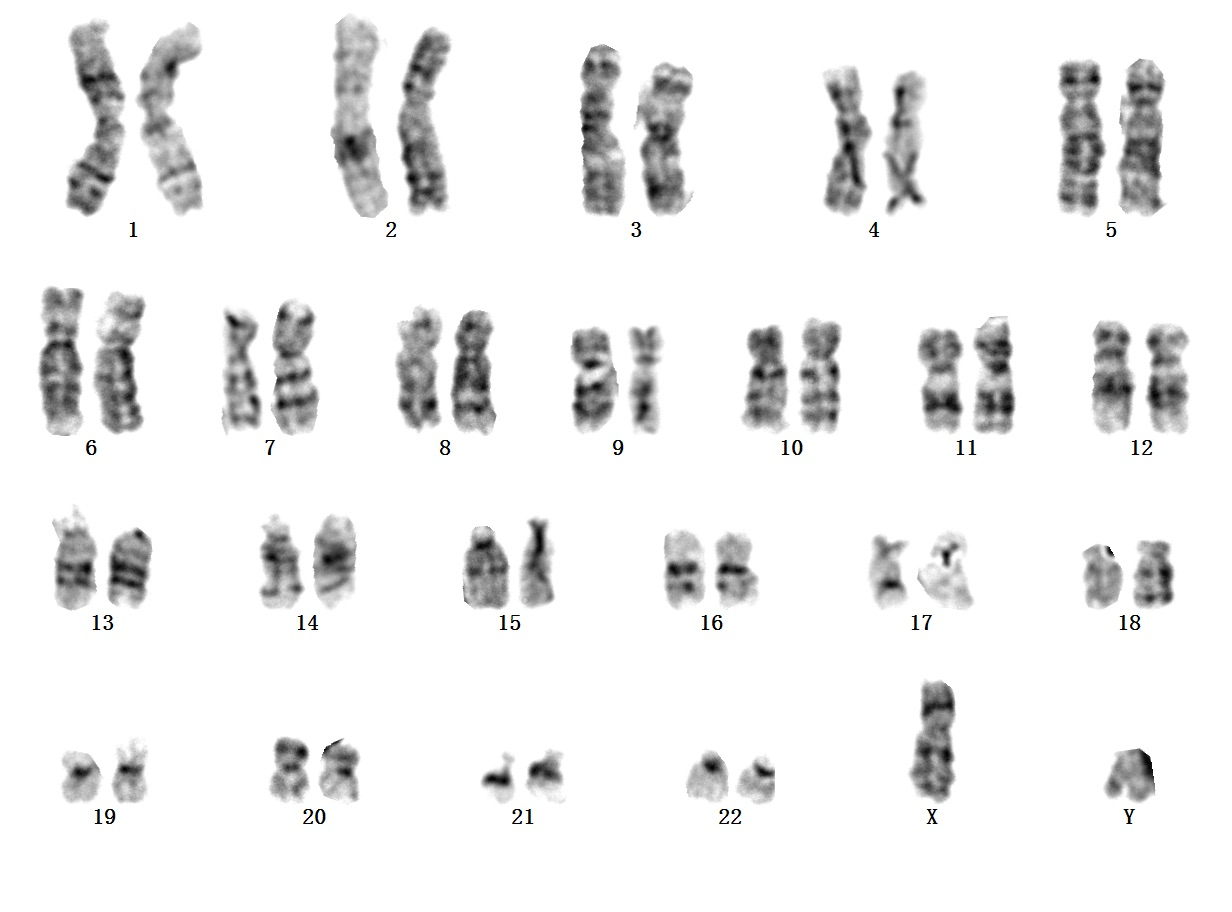}
\caption{\label{Figure-karyo} Karyotype of human chromosomes consisting of 22 autosomes pairs and a pair of sex chromosomes.}
\end{figure}

With the advance in modern image acquisition techniques, digital images of chromosomes become fundamental to the construction of karyotypes (Fig. \ref{Figure-karyo}) and cytogenetic maps for studying structural features \cite{artemov2018development}. Because such tasks are labor-intensive and time-consuming, developing an automatic computer-assisted system has attracted significant research interest for the last 30 years. However, the condensed chromosomes are non-rigid with randomly varying degrees of curvatures along their lengths (Fig. \ref{Figure-karyo}). Such morphological features increase the difficulty of banding pattern analysis and abnormality identification.

\begin{figure*}[htbp]
\centering
\includegraphics[width=6.5in]{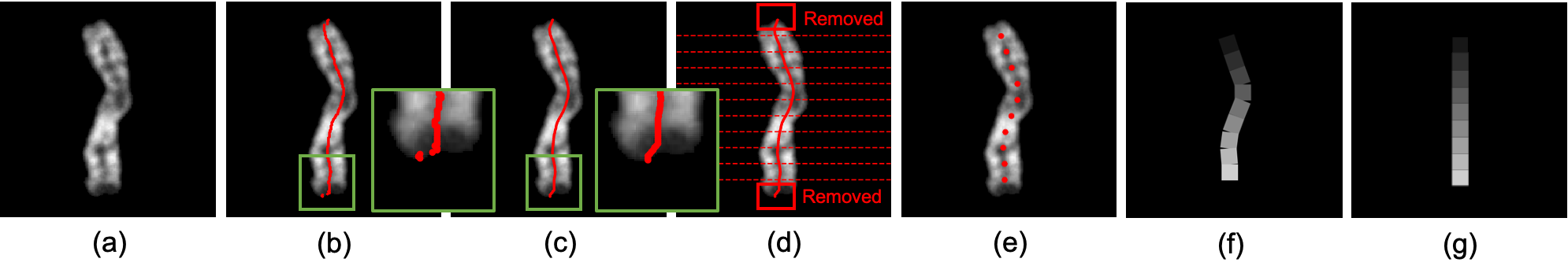}
\caption{\label{Figure-cb} Seven types of images utilized in internal backbone extraction. (a) An example of original chromosomes; (b) an approximate central axis; (c) the smoothed central axis; (d) the smoothed central axis divided into 11 parts; (e) 10-point central axis; (f) the internal backbone; (g) the straightened internal backbone with the same length.}
\end{figure*}

An automatic karyotype construction system typically consists several steps, chromosome segmentation, straightening, classification and arrangement \cite{somasundaram2014straightening,sharma2017crowdsourcing,pardo2018semantic,song2021new,zhang2021chromosome}. Straightened chromosomes have a higher accuracy of chromosome type classification \cite{sharma2017crowdsourcing} and they are pivotal in the development of cytogenetic maps \cite{artemov2018development}. The study of chromosome straightening first begins with cutting paper-based curved chromosome photo into pieces and arranging them into a straightened chromosome \cite{stegniui1991systemic,stegnii1978chromosome}. To the best of our knowledge, based on digital images, current straightening approaches mainly utilize geometric algorithms which are broadly categorized by two approaches: (i) medial axis extraction and (ii) bending points localization. For the first approach, Barrett \textit{et al.} \cite{barrett2003software} requires user interaction and manual labels. References \cite{arora2017algorithm,jahani2012centromere,somasundaram2014straightening} utilize thinning algorithms, such as morphological thinning \cite{guo1989parallel} and Stentiford thinning \cite{stentiford1983some}. However, such algorithms are not suitable for chromosomes with pronounced widths, resulting in many branches along their central axes when thinned \cite{jahani2012centromere,somasundaram2014straightening}. Additionally, when chromosome features are mapped or projected along straightened central axes, the jagged edges remain. The second approach involves analyzing bending points. For straightening, the chromosome is segmented by a single horizontal line from the potential bending point and its two arms are stitched in the vertical direction \cite{roshtkhari2008novel}. Sharma \textit{et al.} \cite{sharma2017crowdsourcing} proposes an improved straightening method based on \cite{roshtkhari2008novel}. It fills the empty region between stitched arms by the mean pixel value at the same horizontal level as reconstructed banding patterns between stitched arms. However, this approach is also not suitable for the chromosomes whose arms are morphologically non-rigid, since the banding patterns of stitched arms are actually rotated rather than straightened along their central axes. Thus the reconstructed chromosomes contain distinct fragments with interrupted banding patterns, and the filled mean pixel value cannot restore realistic banding patterns. Moreover, it has poor performance with misidentifying bending points when there is more than one bending point in a chromosome.

To address the flaws in the geometric algorithms, we propose a novel framework based on image-to-image translation for synthesizing straightened chromosomes with preserved edges and unbroken banding patterns. Furthermore, we are the first to utilize deep learning and generative adversarial networks for straightening chromosomes.

Many studies have shown the success of image-to-image translation in diverse domains, examples including semantic segmentation \cite{ronneberger2015u}, photo generation \cite{yoo2016pixel}, and motion transfer \cite{aberman2019deep,liu2019neural,chan2019everybody}. U-Net \cite{ronneberger2015u} is one of the most popular and effective architectures. Its symmetrical contracting-expanding path structure and skip-connections are pivotal in the preservation of features. Its U-shape architecture has been modified for applications in many studies, such as a hybrid densely connected U-Net \cite{li2018h} and an architecture enhanced by multi-scale feature fusion \cite{dong2020multi}. Pix2pix is a milestone which boosts the performance of conditional generative adversarial networks (cGAN) based on image-to-image translation using a U-shape generator and a patch-wise discriminator \cite{isola2017image}. 

Most applications of image-to-image translation require a large number of paired images. For example, a recent study \cite{chan2019everybody} proposes an effective pipeline for translating human motions by synthesizing target bodies from pose extractions, and it is still trained using large-scale input frames with corresponding pose labels. Based on the mature field of pose detection, the pre-trained state-of-the-art pose detector is used to generate labels from a large number of frames of a given video. Chan \textit{et al.} \cite{chan2019everybody} subsequently trains deep learning models for mapping target body details from each body pose image.

In contrast, it is difficult to acquire sufficient training images and corresponding labels in the research of chromosome straightening. Due to random mutation, structural rearrangement, the non-rigid nature of chromosomes, and different laboratory conditions, it is almost impossible to find two visually identical chromosomes with the same curvature and dyeing condition under microscopes. 

The challenge in this work is to straighten a curved chromosome using only a single chromosome image. Therefore, we propose a novel approach to first extract the internal backbone of the curved chromosome and subsequently increase the size of the chromosome dataset by random image augmentation. Instead of keypoint-based labels, we utilize stick figures as backbones which can retain more augmentation information. The other challenge of this research is to design a model that is able to render realistic and continued chromosome details. At the same time, the straightening algorithm should not be affected by the non-rigid feature of chromosomes. Motivated by this, we innovatively apply image-to-image translation models to learn mapping dependencies from augmented internal backbones to corresponding chromosomes, resulting in high-quality outputs with preserved chromosome details. We also observe that the optimal generator of image-to-image translation models can complement banding patterns and edge details along with given internal backbones. Thus a straightened chromosome is synthesized when we feed a vertical backbone.

The key contributions of this research are three-fold. First, to address the deficiency of inputs, we propose a pertinent augmentation approach to increase the variability of curvatures from the given chromosome and corresponding label simultaneously. Second, using the augmented dataset, we apply two effective image-to-image translation architectures, U-shape networks and cGANs (pix2pix), which demonstrate the efficacy and robustness of our straightening framework. Third, in terms of the accuracy of chromosome type classification, we demonstrate that chromosomes straightened using our framework actually outperform the original curved chromosomes and the ones straightened using geometric algorithms.

The rest of this paper is organized as follows. In Section II, the methodology is described in detail. In Section III, we introduce the data preparation process and illustrate the comparison of straightening results. In Section IV, we discuss the limitations of the proposed approach and present some future research. Finally, we conclude our work in Section V.


\section{Methodology}

In this section, we shall provide a detailed account of our framework. In Section II. A, we propose an approach to generate augmented images and internal backbones from a single curved chromosome. In Section II. B, we describe how the curved chromosome can be straightened by means of its backbone.

\subsection{Data Augmentation Using a Single Image}
For our framework, we propose a two-step strategy to construct an augmented dataset using only one curved chromosome image. 

\begin{algorithm}[h] 
\caption{Chromosome internal backbone Extraction} 
\label{alg::conjugateGradient} 
\hspace*{0.02in} {\bf Input:}
The digital image of a chromosome ($C$) whose width and height are $W$ and $H$, respectively. The background of the image is black (0 pixel values).\\
\hspace*{0.02in} {\bf Output:}
The internal backbone of the chromosome.
\begin{algorithmic}[1] 
\For{each $h\in \{1, 2, ..., H\}$} 
	\If{the current row contains positive pixel values}
		\State find the first ($w_1$) and the last ($w_2$) positions whose pixel value is greater than $0$;
		\State compute the central point $w_c^h = \frac{w_1^h+w_2^h}{2}$;
		\State record the $y$ axis values of the first and the last rows containing positive pixel values as $h_1$ and $h_2$, respectively.
	\EndIf
\EndFor
\State connect all $w_c^h$ to form an approximate central axis extending from $h_1$ to $h_2$;
\State smooth all $w_c^h$ by a moving average algorithm (11-pixel window length), to obtain $w_c^{'h}$;
\State divide the smoothed $w_c^{'h}$ equally into 11 parts (i.e. 12 points) by $y$ axis values in the range of $h_1$ to $h_2$;
\State remove the first and the last parts to obtain a 10-point central axis; 
\State connect the adjacent splitting points by 33-pixel width sticks to obtain a 9-stick internal backbone;
\State generate a vertical 9-stick internal backbone with the same length between the the adjacent splitting points from Line 11.
\end{algorithmic} 
\end{algorithm}

\begin{figure}
\centering
\includegraphics[width=3.5in]{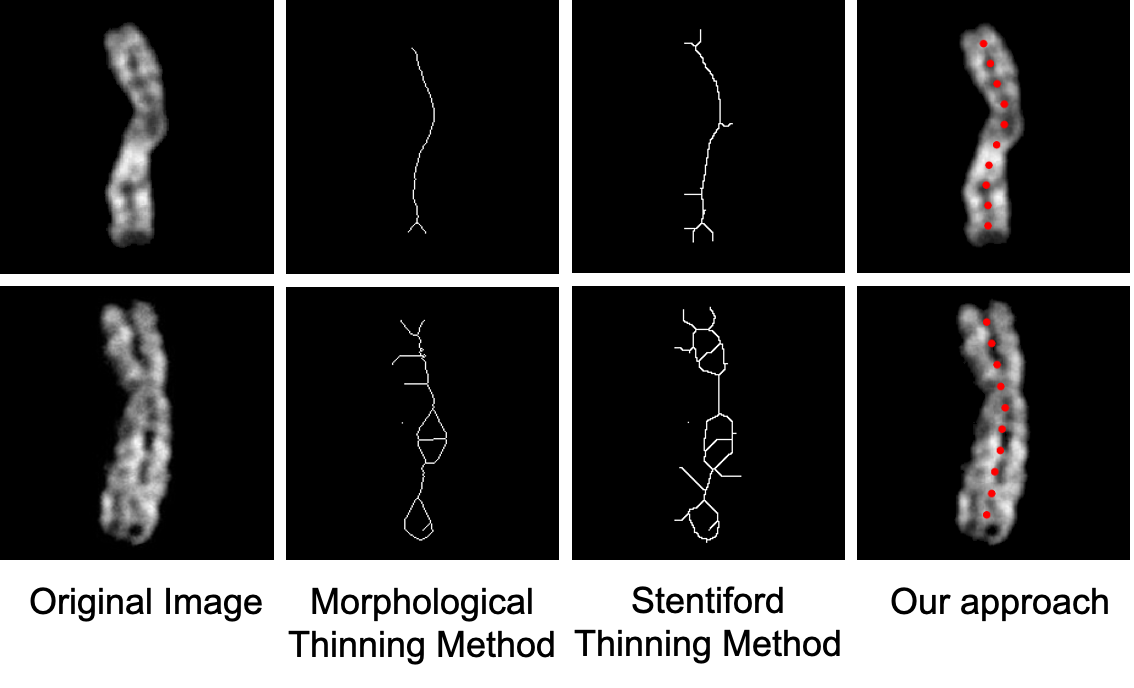}
\caption{\label{Figure-thin} Examples of central axis extraction generated by thinning methods and our approach.}
\end{figure}

\textbf{Step 1.} We construct the label of a curved chromosome (Fig. \ref{Figure-cb}(a)) by extracting a pertinent internal backbone. The entire process is summarized in Algorithm \ref{alg::conjugateGradient}. Considering the chromosome image to be comprised of rows of pixels, the centers of each row are connected to form an approximate central axis extending from top to bottom (Lines 1 to 8 of Algorithm \ref{alg::conjugateGradient}, Fig. \ref{Figure-cb}(b)). To alleviate small-scale fluctuations generated in Line 8, this central axis is then smoothed by a moving average algorithm with an 11-pixel window length \cite{2020SciPy-NMeth} (Line 9, Fig. \ref{Figure-cb}(c)). We divide this smoothed central axis equally into 11 parts in the $y$ axis. Since the first and the last parts may not be aligned in the same directions with both sides of the chromosome (red boxes), these two parts are subsequently removed (Lines 10 to 11, Fig. \ref{Figure-cb}(d) to (e)). The remaining splitting points are connected by 33-pixel width sticks, and these 9 sticks are filled with pixel values in series of equal difference (23, 46, 69, 92, 115, 138, 161, 184, and 207) (Line 12, Fig. \ref{Figure-cb}(f)). This stick figure contains the information of curvature, length, and orientation of the original chromosome. Finally, a vertical backbone is constructed with the same length of each stick (Line 13, Fig. \ref{Figure-cb}(g)), and is fed into the fine-tuned image-to-image translation model for synthesizing the straightened chromosome. 

Fig. \ref{Figure-thin} illustrates that the morphological and Stentiford thinning algorithms may cause branches and irregular rings when the chromosome features pronounced widths. Thus the previous work directed at chromosome straightening \cite{arora2017algorithm,jahani2012centromere,somasundaram2014straightening}, composed of these thinning algorithms, cannot be utilized here. In contrast, our predicted 10-point central axis are approximately in accordance with the actual chromosome backbone.

\begin{figure}
\centering
\includegraphics[width=3.5in]{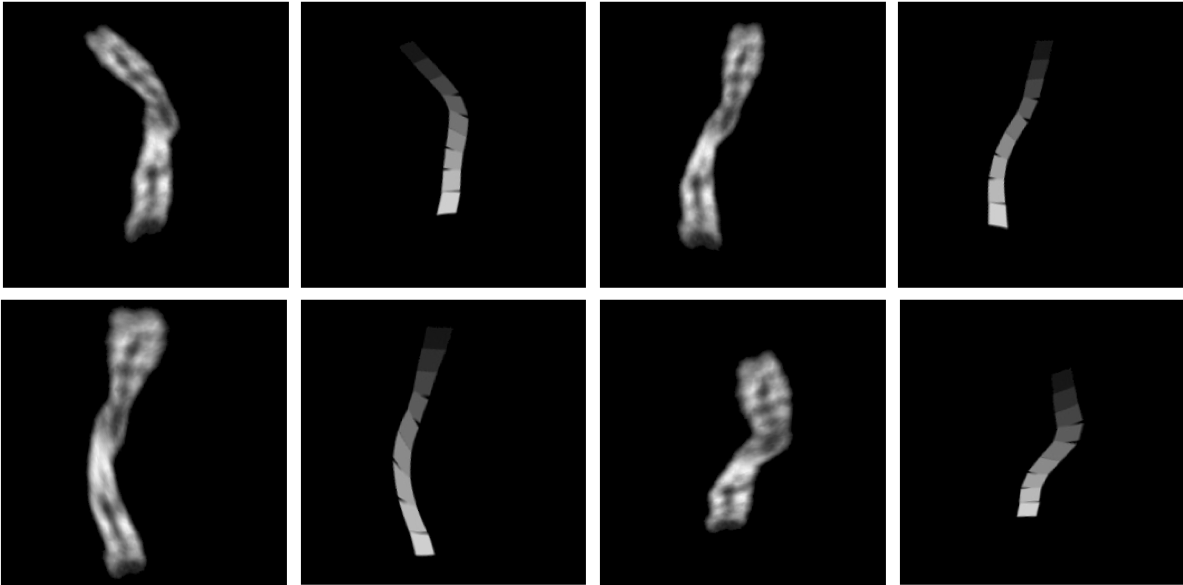}
\caption{\label{Figure-aug} Examples of random data augmentation of a chromosome and corresponding internal backbone.}
\end{figure}

\begin{figure*}
\centering
\subfloat[]{
\begin{minipage}[t]{0.95\linewidth}
\centering
\includegraphics[width=5in]{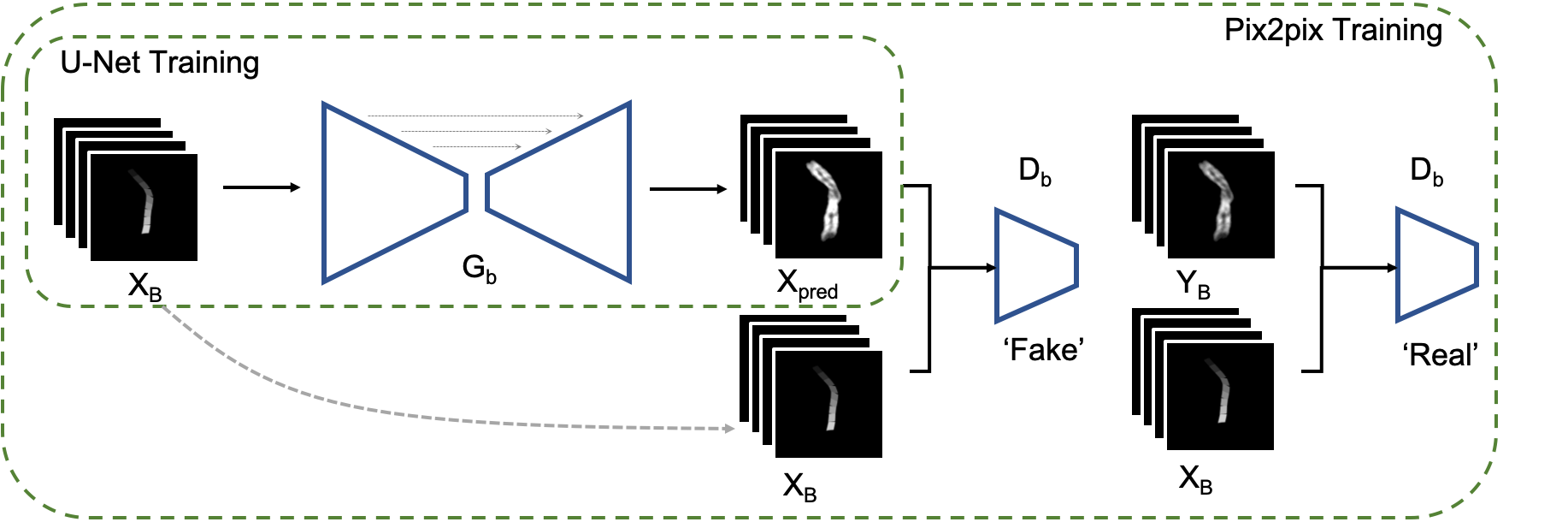}
\end{minipage}%
}%
\par
\subfloat[]{
\begin{minipage}[t]{1.0\linewidth}
\centering
\includegraphics[width=3in]{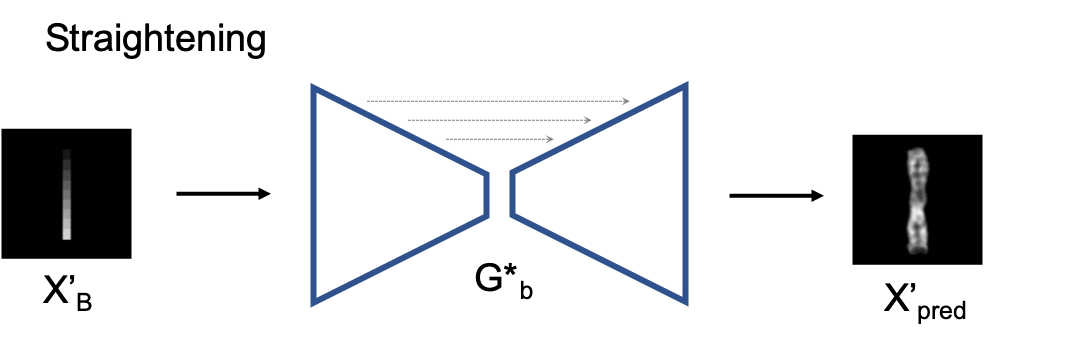}
\end{minipage}%
}%
\caption{The overall process of the proposed framework for chromosome straightening. (a) The training processes of pix2pix or U-Net (the generator part of pix2pix), where $X_B, Y_B$ are augmented backbones and chromosomes and $B \in \{1,...,K\}$ where $K$ is the number of augmented image pairs; $X_{pred}$ is the predicted chromosome image through the generator, $G_b$. (b) The straightening process achieved by the optimal U-Net or generator $G_b^*$. $X_{B}^{'}$ and $X_{pred}^{'}$ are the vertical backbone and the straightened chromosome, respectively.} \label{fig_arch}
\end{figure*}

\textbf{Step 2.} We improve the performance of deep learning models by generating more augmented chromosomes with different degrees of curvatures. We first apply random elastic deformation \cite{gvtulder2019} and random rotation (from -45 to 45 degree) to the curved chromosome and its backbone simultaneously (Fig. \ref{Figure-cb}(a) and (f)) until a sizeable number of augmented chromosomes and backbones (1000 pairs in this research) are obtained for training and validation (Fig. \ref{Figure-aug}). Note that the setup of the elastic deformation algorithm \cite{gvtulder2019} is $points = 3$ and $sigma = 18$ for $256\times256$ images, in order to generate plausible virtual curvatures. Since we utilize 33-pixel width sticks, rather than key points to label internal backbones, the detailed augmentation information, such as stretching, rotation and distortion, is retained and learned by the image-to-image translation models.

\subsection{Image-to-Image Translation for Straightening}

Since the objective of this study is to input a straightened backbone of a chromosome for synthesizing the corresponding chromosomes with preserved banding patterns, our novel image-to-image translation models are object specific. Therefore, it is essential to construct an augmented dataset for each image-to-image translation model. Utilizing the approach mentioned in Step 2, we generate 1000 augmented image pairs for each curved chromosome. The augmented dataset is then randomly split using a ratio of 9:1 for training and validation, respectively. Under our framework, we shall utilize two image-to-image translation models, U-Net and pix2pix (Fig. \ref{fig_arch}(a)). It should be noted that the U-Net utilized in this research is identical to the generator part of pix2pix. The training process of U-Net is a regular supervised learning method achieved by synthesized chromosomes and corresponding ground-truths. In pix2pix, a generator $G_b$ synthesizes chromosomes from the augmented backbones to mislead $D_b$. Meanwhile, a discriminator $D_b$ is trained for discerning ``real" images from ``fake" images yielded by the generator. The $G_b$ and $D_b$ is optimized with the objective function:
\begin{equation}
G_b^* = \arg \min_{G_b} \max_{D_b} \mathcal{L}_{cGAN}(G_b,D_b) + \lambda\mathcal{L}_{pix}(G_b)
\end{equation}
where $G_b^*$ represents the optimal generator; $\lambda$ is a coefficient to balance two losses; $\mathcal{L}_{cGAN}(G_b,D_b)$ is the adversarial loss (Equation 2); and $\mathcal{L}_{pix}(G_b)$ is L1 distance to evaluate pixel-wise performance between generated images and ground-truths (Equation 3): 

\begin{equation}
\begin{aligned}
\mathcal{L}_{cGAN}(G_b,D_b) = \mathbb{E}_{x_B,z}[(D_b(x_B,G_b(x_B,z))-1)^2] + \\
\mathbb{E}_{x_B,y_B}[(D_b(x_B,y_B))^2]
\end{aligned}
\end{equation}

\begin{equation}
\mathcal{L}_{pix}(G_b) = \mathbb{E}_{x_B,y_B,z}[\|y_B-G(x_B,z)\|_{1}]
\end{equation}
In the above: $x_B$ and $y_B$ represent augmented backbones and chromosomes, respectively; $B \in \{1,...,K\}$ where $K$ is the number of augmented pairs that we want; and $z$ is the noise introduced in the generator. 

To straighten the chromosome, we input its vertical backbone (Fig. \ref{Figure-cb}(g)) into the optimal U-Net or optimal generator $G_b^*$, which will output the corresponding chromosome (Fig. \ref{fig_arch}(b)). 


\section{Experiments and Results}

\subsection{Chromosome Dataset}

To test our framework on real-world images, we extract 642 low-resolution human chromosome images from karyotypes provided by a biomedical company. Images in this research have been cleaned so that connections between these images and their corresponding owners have been removed. Since the chromosomes with relatively long arms and noticeable curvatures require straightening (Figure \ref{Figure-karyo}), we collect Type 1 to 7 chromosomes in this research. We invert the color of these grey-scale images and center them in a $256\times256$ black background. As described in Section II. A, 1000 augmented image pairs were obtained from each curved chromosome image before feeding into the U-Net and pix2pix models. It should be noted here that each augmented dataset is individually trained for straightening since our framework is object specific. 

\subsection{Evaluation Metrics}

We apply two evaluation metrics to quantitatively measure the performance of these straightening methods. Due to the obvious morphological deformation between straightened results and original curved chromosomes, traditional similarity measurement metrics, such as Euclidean distance, structural similarity index (SSIM) \cite{wang2004image} and peak-signal-to-noise ratio (PSNR) \cite{hore2010image}, designed for evaluating image quality degradation generated by image processing or compression, are not suitable for this task. Instead, Learned Perceptual Image Patch Similarity (LPIPS) \cite{zhang2018unreasonable} was used to evaluate straightening performance of different methods in this paper. The LPIPS is an emergent deep neural network-based method which is able to extract deep features of images for evaluating high-order structure similarity. Compared to the results of these traditional metrics, its results are more in accordance with human perceptual similarity judgment \cite{zhang2018unreasonable}.

Apart from LPIPS, to ensure the details of straightened results are preserved in practice, we also assess the effectiveness of different straightening methods based on chromosome type classification. If the banding patterns and edge details of chromosomes are well preserved during straightening, the classification accuracy of straightened chromosomes should not decrease. In contrast, unpreserved details, such as broken bands, may not provide enough information for the classification model. The original images (642 curved chromosomes, Type 1 to 7) are randomly split using the ratio of 3:1 for 4-fold cross-validation. With a fixed random seed, this process is similarly carried out for the straightened chromosomes generated by different methods.

\begin{figure*}
\centering
\includegraphics[width=4.4in]{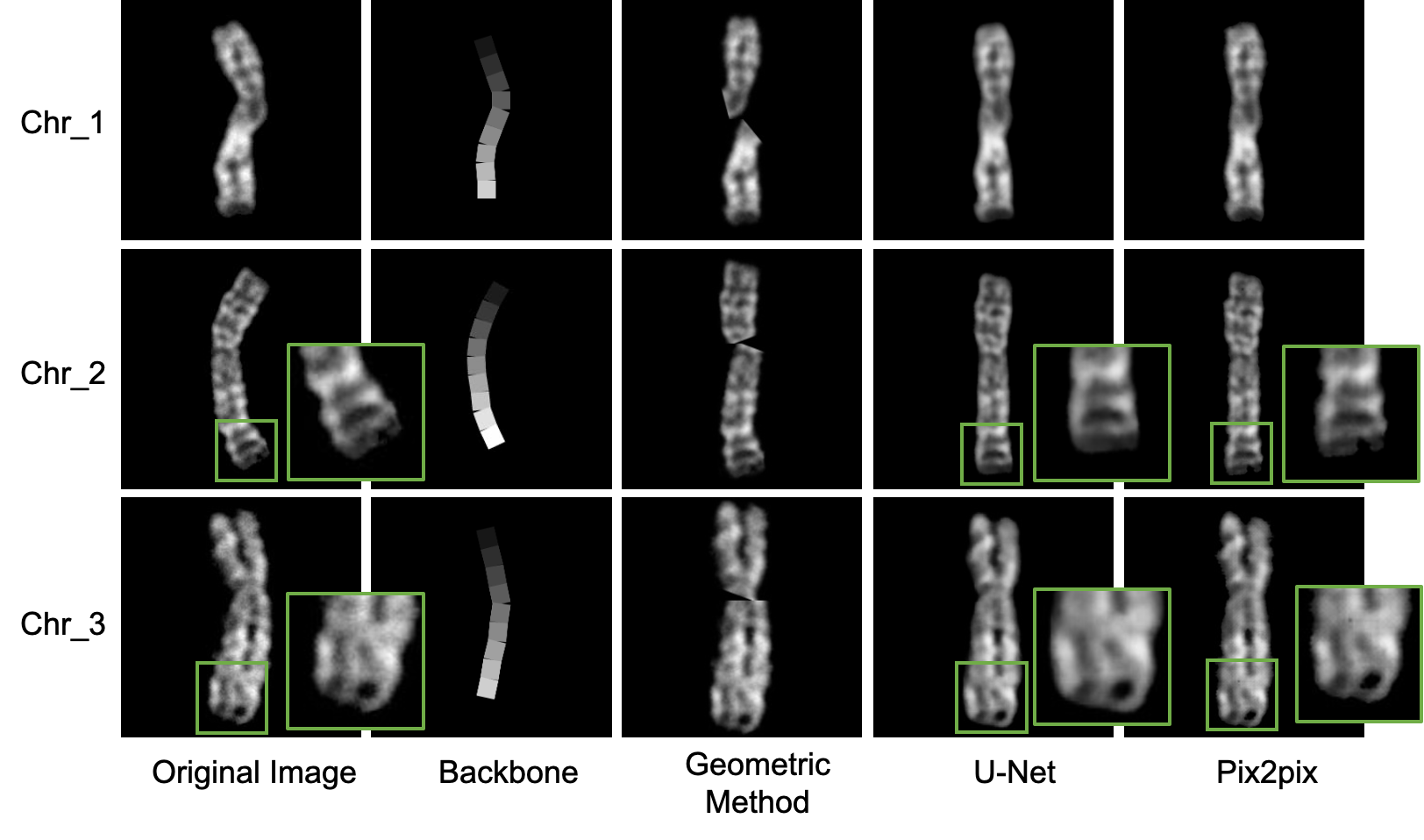}
\caption{Three examples of straightening results. From left to right: original images, the geometric method \cite{roshtkhari2008novel,sharma2017crowdsourcing}, our framework using U-Net and pix2pix. Enlarged regions demonstrate marginally improved details of pix2pix over U-Net.} \label{fig_exam}
\end{figure*}

\subsection{Implementation Details}

\begin{table*}[htbp]
\caption{LPIPS results on different chromosome datasets (mean $\pm$ std.). For LPIPS lower is more similar.}\label{table-lpips}
\centering
\begin{tabular}{|l |c |c |c |c |}
\hline
{\bfseries } & Original Images \textit{vs.} Geometric Method & Original Images \textit{vs.} U-Net & Original Images \textit{vs.} Pix2pix & U-Net \textit{vs.} Pix2pix \\
\hline
LPIPS & $0.1621 \pm 0.052$ & $ \textbf{0.1356} \pm \textbf{0.051}$ & $ \textbf{0.1318} \pm \textbf{0.050}$ & $\textbf{0.0239} \pm \textbf{0.011}$ \\
\hline
\end{tabular}
\end{table*}

Our experiments are implemented using PyTorch and run on two NVIDIA RTX 2080Ti GPUs. In each training process of chromosome straightening, the training and validation sets are split by a fixed random seed. The input image pairs are first normalized by default values (mean $\mu =$ 0.5 and standard deviation $\sigma =$ 0.5), and these results are fed into image-to-image translation models for learning the mapping dependence from backbones to chromosomes. Models are trained with an initial learning rate $lr = 0.00004$. The validation performance is checked three times per epoch, and the weights are saved when the best validation performance is updated. When the validation performance does not improve for 9 consecutive checks, the learning rate is reduced to 80\% for fine-tuning. To avoid overfitting, the training process is terminated when there are 27 consecutive checks without updated validation performance. For each chromosome type classification model (Alexnet \cite{krizhevsky2012imagenet}, ResNet50 \cite{he2016deep} and DenseNet169 \cite{huang2017densely}), the training process is initialized with a learning rate of $lr = 0.00004$ and corresponding ImageNet pre-trained weights. We utilize 12 and 120 consecutive checks for fine-tuning and avoiding overfitting, respectively. Furthermore, we use identical random seeds, preprocessing and hyperparameter settings for 4-fold cross-validation of the chromosome type classification.


\subsection{Results}

\begin{figure*}
\centering
\includegraphics[width=6.7in]{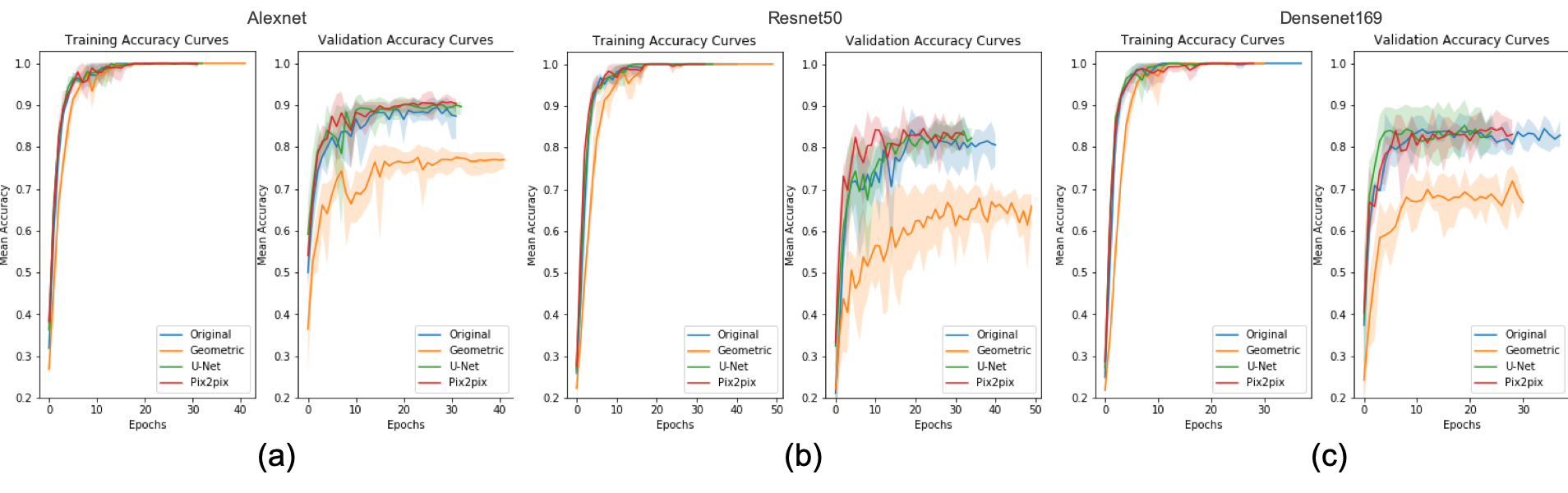}
\caption{\label{Figure-training} Training and validation accuracy curves of three CNN models for chromosome type classification (4-fold cross-validation). Shadow regions represent the range over four folds and solid lines represent mean accuracy.}
\end{figure*}

\subsubsection{Comparison of Straightening Performance}

Although there are two categories of geometric methods (medial axis extraction \cite{arora2017algorithm,jahani2012centromere,somasundaram2014straightening} and bending points localization \cite{roshtkhari2008novel,sharma2017crowdsourcing}), we found that the morphological and Stentiford thinning algorithms of medial axis extraction may cause many unexpected branches and irregular rings. Therefore, we investigated the performance of chromosome straightening using: (a) the geometric method (bending points localization) whose main component is used by \cite{roshtkhari2008novel,sharma2017crowdsourcing}, and our image-to-image translation model based framework with (b) U-Net and (c) pix2pix models.

Fig. \ref{fig_exam} gives three examples of the straightening results using the 642 curved chromosomes. The five columns correspond to: (i) the original unstraightened images, (ii) corresponding backbones extracted by our approach, (iii) outputs of the geometric method \cite{roshtkhari2008novel,sharma2017crowdsourcing}, as well as the results from our framework with (iv) U-Net and (v) pix2pix, respectively. Although \cite{sharma2017crowdsourcing} additionally fills empty regions between stitched arms with the mean pixel values at the same horizontal level, the main problem of \cite{roshtkhari2008novel} whose results contain distinct segmented banding patterns between arms is still unresolved. In the third column of Fig. \ref{fig_exam}, we illustrate results of the straightening algorithm whose key part is used in \cite{roshtkhari2008novel,sharma2017crowdsourcing}. As examples in the third column of Chr\_1 and Chr\_2, the performance of the geometric method further deteriorates if there are curved arms and more than one bending point. Compared to these results, our framework demonstrates superiority both in translation consistency and in non-rigid straightening results (the fourth and fifth columns). The curvature of arms and the number of bending points do not decrease the performance of our framework because the image-to-image translation based framework relies on backbones rather than through morphological analysis. Since the provided chromosomes are low-resolution images, we notice that some straightened chromosomes (e.g. Chr\_1) of U-Net and pix2pix have indistinguishable synthesized internal details and intensity. For many examples (enlarged area in Fig. \ref{fig_exam}), pix2pix marginally outperforms the U-Net model with more preserved edge details achieved by the patch-wise discriminator and adversarial training method. Since the chromosome images in this research are low-resolution ($256 \times 256$), the ability to generate fine details using our framework with cGANs may become more obvious in high-resolution chromosome straightening and could be extended for use in the development of cytogenetic maps. 

The average values and standard deviations (std.) of LPIPS are summarized in Table \ref{table-lpips}. Since LPIPS shows the perceptual distance between two images even there is obvious deformation, we quantify the similarity between curved chromosomes and straightened ones. We can observe that the straightening results of the pix2pix model under our framework achieves the best performance with a minimum LPIPS value (the third column of Table \ref{table-lpips}). The measurement of Original Images \textit{vs.} U-Net and U-Net \textit{vs.} Pix2pix indicates that the performance of U-Net is slightly worse than pix2pix due to the superior translation consistency of cGANs to U-shape neural networks. As a comparison, straightening results of the geometric method produced the highest LPIPS value, which may be caused by the broken banding patterns between stitched arms.

\begin{figure*}
\centering
\includegraphics[width=4.4in]{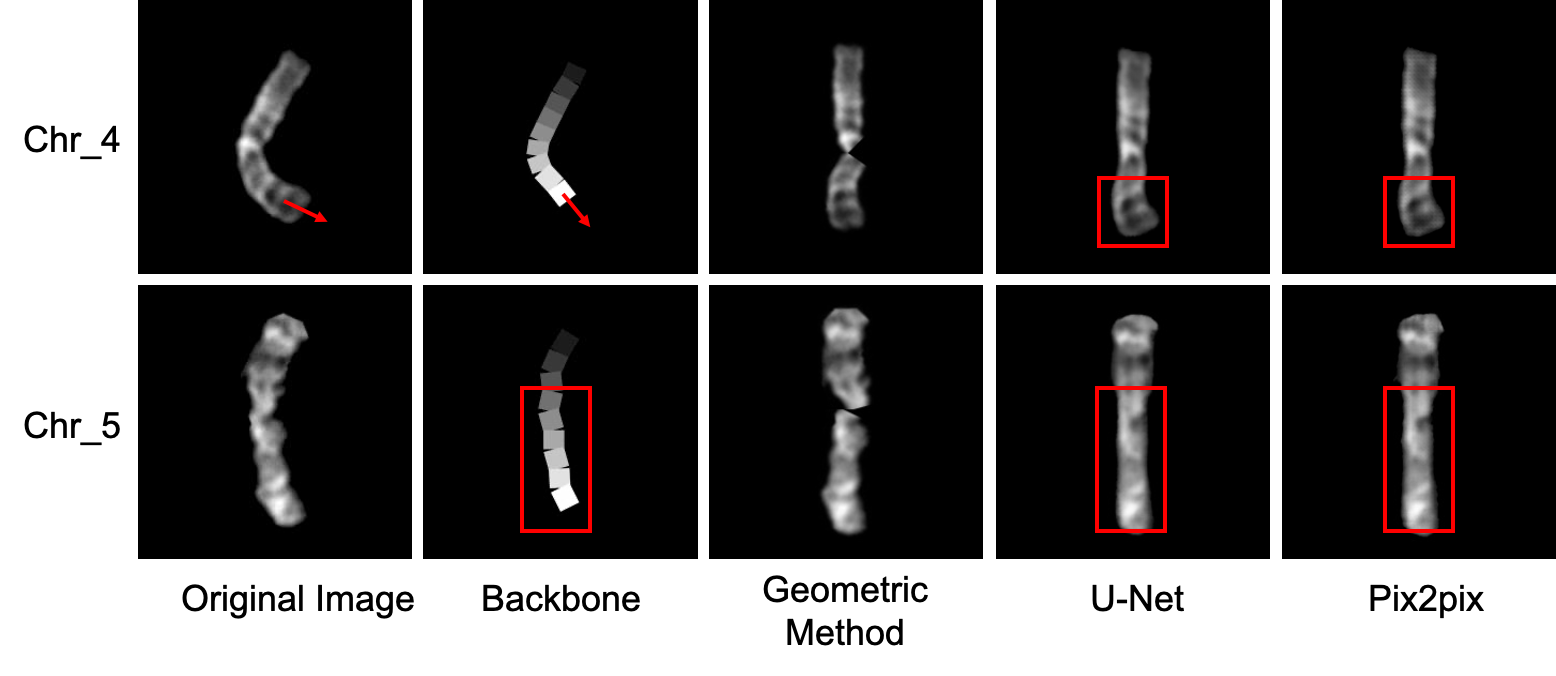}
\caption{Two examples of failure cases. From left to right: original images, the geometric method \cite{roshtkhari2008novel,sharma2017crowdsourcing}, our framework using U-Net and pix2pix.} \label{fig_fail}
\end{figure*}

\subsubsection{Comparison of Chromosome Type Classification Results on Different Straightened Datasets}

We also performed experiments to determine if our proposed straightening framework enhanced the accuracy of the chromosome type classification. It is significant because the assessment of classification accuracy is an indispensable step in automatic karyotyping analysis \cite{sharma2017crowdsourcing,zhang2018chromosome,qin2019varifocal}. Inaccurate straightened results may obscure the unique morphological features and banding patterns of different chromosome types.

\begin{table}[htbp]
\caption{Comparison of averaged classification accuracy (4-fold cross-validation)}\label{tab2}
\centering
\begin{tabular}{|l |c |c |c |l |}
\hline
{\bfseries Accuracy (\%)} &Alexnet & ResNet50 &DenseNet169\\
\hline
Original Images (Baselines) & 90.47 & 85.31 & 86.09\\
Geometric Method \cite{roshtkhari2008novel,sharma2017crowdsourcing} & 78.44 & 70.16 & 73.59\\
U-Net & \textbf{91.51} & \textbf{85.65} & \textbf{87.65} \\
Pix2pix & \textbf{91.67} & \textbf{86.57} & \textbf{87.81}\\
\hline
\end{tabular}
\end{table}

\begin{table}[htbp]
\caption{Comparison of averaged AUC of chromosome type classification (4-fold cross-validation)}\label{tab3}
\centering
\begin{tabular}{|l |c |c |c |l |}
\hline
{\bfseries AUC} &Alexnet & ResNet50 &DenseNet169\\
\hline
Original Images (Baselines) & 0.9423 & 0.9163 & 0.9271\\
Geometric Method \cite{roshtkhari2008novel,sharma2017crowdsourcing} & 0.8513 & 0.8317 & 0.8513\\
U-Net & \textbf{0.9487} & \textbf{0.9204} & \textbf{0.9301} \\
Pix2pix & \textbf{0.9510} & \textbf{0.9293} & \textbf{0.9311}\\
\hline
\end{tabular}
\end{table}

\begin{figure*}
\centering
\includegraphics[width=6in]{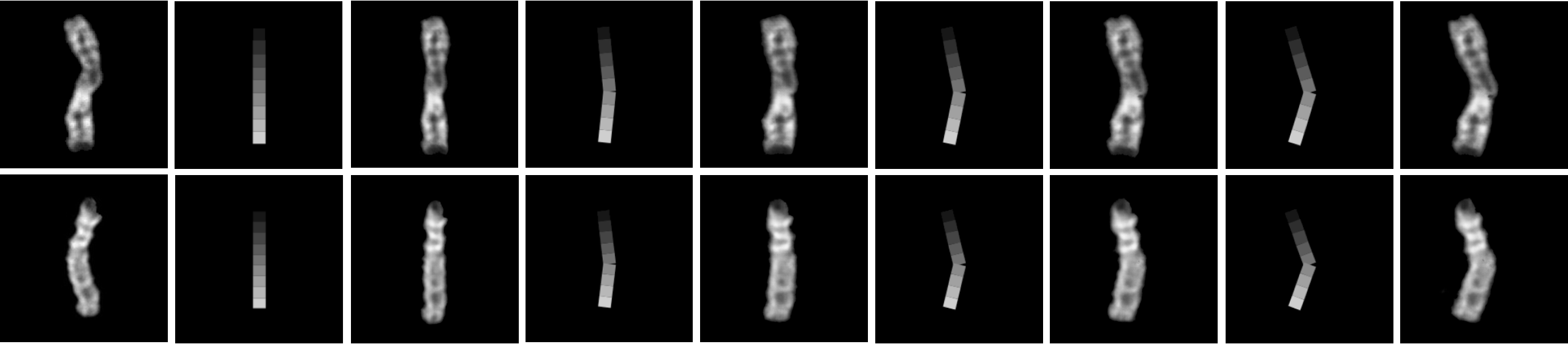}
\caption{Examples of synthesized results with a series of curved internal backbones (generated by our framework with the pix2pix model).} \label{fig_multi}
\end{figure*}

Tables \ref{tab2} and \ref{tab3} give the comparisons between three standard state-of-the-art classification networks, AlexNet \cite{krizhevsky2012imagenet}, ResNet50 \cite{he2016deep} and DenseNet169 \cite{huang2017densely}. The accuracy scores and their Area Under Curve (AUC) are the mean value of 4-fold cross-validation results. We consider the scores trained by original curved chromosomes as baselines. We can see that wrongly identified bending points and stitched chromosome arms with discontinued banding patterns from the geometric method, reduce the classification results by a significant margin (-13.23\% accuracy, -0.084 AUC on average). In contrast, our framework achieves top scores and marginally outperforms the baselines by 0.98\% accuracy, 0.0045 AUC (U-Net) and 1.39\% accuracy, 0.0085 mean AUC (pix2pix) on average. One possible reason is that the straightened and uninterrupted banding patterns help neural networks to learn uncurved and unrotated unique features of chromosomes. The superiority of our proposed framework suggests that it may benefit banding pattern identification and abnormality detection in the automatic pathological diagnosis of karyotypes. Fig. \ref{Figure-training} depicts the mean accuracy curves of different training/validation sets of these three models. It illustrates that the chromosome type classification performance of datasets between original images, chromosomes generated by U-Net and pix2pix display similar trends, which is in accordance with the results of Table \ref{tab2} and Table \ref{tab3}. This indicates the details of chromosomes are well preserved after straightening. In contrast, the chromosome type classification accuracy is severely affected by the discontinued banding patterns and unstraightened arms generated by the geometric method.


\section{Limitation and Discussion}

\subsection{Computation Time}

To address the flaws, such as the broken banding patterns in geometric methods and random stretching in elastic deformation algorithms, we propose a chromosome straightening framework which is object specific. Therefore, it is time-consuming to train a separate straightening model for every curved chromosome. In future research, a generalized chromosome straightening model shall be designed. We would design an improved model for disentangling the information of internal backbones and banding patterns.


\subsection{Failure Cases}

Under our framework, we notice two types of failure cases. First, the straightening performance hinges on the accuracy of the central axes identified. When the curvature of a chromosome is too large, the extracted internal backbone may not be aligned in a similar direction with the original image (red arrows of Chr\_4 in Fig. \ref{fig_fail}). In this case, the relation between the backbone and corresponding banding patterns are still preserved. As a result, that part may not be well straightened. Second, some irregular chromosomes may still cause small-scale fluctuations of backbones even after the moving average algorithm, resulting in blurred synthesized banding patterns and edge details (Chr\_5 in Fig. \ref{fig_fail}). Because of this, high-quality labels of chromosomes are still deficient in the augmented dataset. A plausible direction would be an improvement of the backbone extraction method. A crowdsourcing database of labeled backbones could be established for developing a powerful deep learning based backbone detector of chromosomes.

\subsection{Potential Applications}

Since the results of our straightening framework demonstrate a higher classification accuracy, it is worthwhile to incorporate the framework into automatic karyotyping analysis and cytogenetic map construction. With the development of image-to-image translation research, many advanced modules and architectures, for example, attention-based GANs \cite{zhang2018self}, may be integrated into our framework to further improve its efficacy and robustness. 

Since our augmented datasets contain information concerning random deformation and rotation, we observe that fine-tuned generators not only have an ability to straighten chromosomes, but also can synthesize more chromosomes by inputting internal backbones with different curvatures (Fig. \ref{fig_multi}). Therefore, our framework shows the potentiality for generating augmented chromosomes with highly preserved detail along with customized backbone images.

Compared to regular U-shape networks, cGANs have more potential in the application of high-resolution chromosome straightening with higher translation consistency. In the latest study, Artemov \textit{et al.} \cite{artemov2018development} employs PhotoShop for straightening high-resolution chromosomes when developing cytogenetic maps, so an automatic high-resolution chromosome straightening framework is still in demand. Similar to the evolution from pix2pix to pix2pixHD \cite{wang2018high}, our straightening framework may also be further modified for high-resolution chromosome images.

\section{Conclusions}

In this study, we propose a novel image-to-image translation based chromosome straightening framework which sets a new direction for object straightening. The framework transforms the task of straightening into the learning of mapping dependency from randomly augmented backbones to corresponding chromosomes. It allows straightened chromosomes to be generated from vertical backbones. The straightening performance of our framework is significantly better than the geometric approach with more realistic images of uninterrupted banding patterns. Under our framework, the average classification accuracy of U-Net and pix2pix evaluated by state-of-the-art classification models is higher than the baselines by 0.98\% and 1.39\%, respectively. However, using this straightening framework it is still computationally expensive to train separate models for different curved chromosomes, the framework also may generate blurred results due to inaccurately identified internal backbones. Since the study of deep learning based chromosome straightening is at its infancy, many improvements can be made to our framework, such as a more accurate internal backbone extraction method, and a generalized architecture which is not object specific.

\section*{Acknowledgment}

This work was supported in part by the National Science Foundation of China (NSFC) under Grant 61501380, in part by the Key Program Special Fund in XJTLU (KSF-T-01), in part by the Key Program Special Fund in XJTLU (KSF-A-22), in part by the Key Programme Special Fund in XJTLU (KSF-E-21), in part by the Neusoft Corporation (item number SKLSAOP1702), in part by the Shenzhen Science and Technology Innovation Commission (Grants No. JCYJ20170410172100520), and in part by the Shenzhen Institute of Artificial Intelligence and Robotics for Society (Grants No. 2019-INT020).



%



\bibliographystyle{IEEEtran}
\bibliography{Manuscript}

\begin{thebibliography}{10}
\providecommand{\url}[1]{#1}
\csname url@samestyle\endcsname
\providecommand{\newblock}{\relax}
\providecommand{\bibinfo}[2]{#2}
\providecommand{\BIBentrySTDinterwordspacing}{\spaceskip=0pt\relax}
\providecommand{\BIBentryALTinterwordstretchfactor}{4}
\providecommand{\BIBentryALTinterwordspacing}{\spaceskip=\fontdimen2\font plus
\BIBentryALTinterwordstretchfactor\fontdimen3\font minus
  \fontdimen4\font\relax}
\providecommand{\BIBforeignlanguage}[2]{{%
\expandafter\ifx\csname l@#1\endcsname\relax
\typeout{** WARNING: IEEEtran.bst: No hyphenation pattern has been}%
\typeout{** loaded for the language `#1'. Using the pattern for}%
\typeout{** the default language instead.}%
\else
\language=\csname l@#1\endcsname
\fi
#2}}
\providecommand{\BIBdecl}{\relax}
\BIBdecl

\bibitem{speicher2005new}
M.~R. Speicher and N.~P. Carter, ``The new cytogenetics: blurring the
  boundaries with molecular biology,'' \emph{Nature reviews genetics}, vol.~6,
  no.~10, pp. 782--792, 2005.

\bibitem{saidzhafarova2009molecular}
A.~Saidzhafarova, G.~Artemov, T.~Karamysheva, N.~Rubtsov, and V.~Stegnii,
  ``Molecular cytogenetic analysis of dna from pericentric heterochromatin of
  chromosome 2l of malaria mosquito anopheles beklemishevi (culicidae,
  diptera),'' \emph{Russian journal of genetics}, vol.~45, no.~1, pp. 49--53,
  2009.

\bibitem{hills2006cri}
C.~Hills, J.~H. Moller, M.~Finkelstein, J.~Lohr, and L.~Schimmenti, ``Cri du
  chat syndrome and congenital heart disease: a review of previously reported
  cases and presentation of an additional 21 cases from the pediatric cardiac
  care consortium,'' \emph{Pediatrics}, vol. 117, no.~5, pp. e924--e927, 2006.

\bibitem{kostanecka2012developmental}
A.~Kostanecka, L.~B. Close, K.~Izumi, I.~D. Krantz, and M.~Pipan,
  ``Developmental and behavioral characteristics of individuals with
  pallister--killian syndrome,'' \emph{American Journal of Medical Genetics
  Part A}, vol. 158, no.~12, pp. 3018--3025, 2012.

\bibitem{artemov2018development}
G.~N. Artemov, V.~N. Stegniy, M.~V. Sharakhova, and I.~V. Sharakhov, ``The
  development of cytogenetic maps for malaria mosquitoes,'' \emph{Insects},
  vol.~9, no.~3, p. 121, 2018.

\bibitem{somasundaram2014straightening}
D.~Somasundaram and V.~V. Kumar, ``Straightening of highly curved human
  chromosome for cytogenetic analysis,'' \emph{Measurement}, vol.~47, pp.
  880--892, 2014.

\bibitem{sharma2017crowdsourcing}
M.~Sharma, O.~Saha, A.~Sriraman, R.~Hebbalaguppe, L.~Vig, and S.~Karande,
  ``Crowdsourcing for chromosome segmentation and deep classification,'' in
  \emph{Proceedings of the IEEE Conference on Computer Vision and Pattern
  Recognition Workshops}, 2017, pp. 34--41.

\bibitem{pardo2018semantic}
E.~Pardo, J.~M.~T. Morgado, and N.~Malpica, ``Semantic segmentation of mfish
  images using convolutional networks,'' \emph{Cytometry Part A}, vol.~93,
  no.~6, pp. 620--627, 2018.

\bibitem{song2021new}
S.~Song, T.~Bai, Y.~Zhao, W.~Zhang, C.~Yang, J.~Meng, F.~Ma, and J.~Su, ``A new
  convolutional neural network architecture for automatic segmentation of
  overlapping human chromosomes,'' \emph{Neural Processing Letters}, pp. 1--17,
  2021.

\bibitem{zhang2021chromosome}
J.~Zhang, W.~Hu, S.~Li, Y.~Wen, Y.~Bao, H.~Huang, C.~Xu, and D.~Qian,
  ``Chromosome classification and straightening based on an interleaved and
  multi-task network,'' \emph{IEEE Journal of Biomedical and Health
  Informatics}, 2021.

\bibitem{stegniui1991systemic}
Stegni{\u\i}, ``Systemic reorganization of the architechtonics of polytene
  chromosomes in onto-and phylogenesis of malaria mosquitoes. structural
  features regional of chromosomal adhesion to the nuclear membrane.''

\bibitem{stegnii1978chromosome}
V.~Stegnii and V.~Kabanova, ``Chromosome analysis of anopheles atroparvus and
  anopheles maculipennis (diptera, culicidae),'' \emph{Zoologicheskii zhurnal},
  1978.

\bibitem{barrett2003software}
S.~Barrett and C.~De~Carvalho, ``A software tool to straighten curved
  chromosome images,'' \emph{Chromosome Research}, vol.~11, no.~1, pp. 83--88,
  2003.

\bibitem{arora2017algorithm}
T.~Arora, R.~Dhir, and M.~Mahajan, ``An algorithm to straighten the bent human
  chromosomes,'' in \emph{2017 Fourth International Conference on Image
  Information Processing (ICIIP)}.\hskip 1em plus 0.5em minus 0.4em\relax IEEE,
  2017, pp. 1--6.

\bibitem{jahani2012centromere}
S.~Jahani and S.~K. Setarehdan, ``Centromere and length detection in
  artificially straightened highly curved human chromosomes,''
  \emph{International journal of Biological engineering}, vol.~2, no.~5, pp.
  56--61, 2012.

\bibitem{guo1989parallel}
Z.~Guo and R.~W. Hall, ``Parallel thinning with two-subiteration algorithms,''
  \emph{Communications of the ACM}, vol.~32, no.~3, pp. 359--373, 1989.

\bibitem{stentiford1983some}
F.~Stentiford and R.~Mortimer, ``Some new heuristics for thinning binary
  handprinted characters for ocr,'' \emph{IEEE transactions on systems, man,
  and cybernetics}, no.~1, pp. 81--84, 1983.

\bibitem{roshtkhari2008novel}
M.~J. Roshtkhari and S.~K. Setarehdan, ``A novel algorithm for straightening
  highly curved images of human chromosome,'' \emph{Pattern recognition
  letters}, vol.~29, no.~9, pp. 1208--1217, 2008.

\bibitem{ronneberger2015u}
O.~Ronneberger, P.~Fischer, and T.~Brox, ``U-net: Convolutional networks for
  biomedical image segmentation,'' in \emph{International Conference on Medical
  image computing and computer-assisted intervention}.\hskip 1em plus 0.5em
  minus 0.4em\relax Springer, 2015, pp. 234--241.

\bibitem{yoo2016pixel}
D.~Yoo, N.~Kim, S.~Park, A.~S. Paek, and I.~S. Kweon, ``Pixel-level domain
  transfer,'' in \emph{European Conference on Computer Vision}.\hskip 1em plus
  0.5em minus 0.4em\relax Springer, 2016, pp. 517--532.

\bibitem{aberman2019deep}
K.~Aberman, M.~Shi, J.~Liao, D.~Lischinski, B.~Chen, and D.~Cohen-Or, ``Deep
  video-based performance cloning,'' in \emph{Computer Graphics Forum},
  vol.~38, no.~2.\hskip 1em plus 0.5em minus 0.4em\relax Wiley Online Library,
  2019, pp. 219--233.

\bibitem{liu2019neural}
L.~Liu, W.~Xu, M.~Zollhoefer, H.~Kim, F.~Bernard, M.~Habermann, W.~Wang, and
  C.~Theobalt, ``Neural rendering and reenactment of human actor videos,''
  \emph{ACM Transactions on Graphics (TOG)}, vol.~38, no.~5, pp. 1--14, 2019.

\bibitem{chan2019everybody}
C.~Chan, S.~Ginosar, T.~Zhou, and A.~A. Efros, ``Everybody dance now,'' in
  \emph{Proceedings of the IEEE International Conference on Computer Vision},
  2019, pp. 5933--5942.

\bibitem{li2018h}
X.~Li, H.~Chen, X.~Qi, Q.~Dou, C.-W. Fu, and P.-A. Heng, ``H-denseunet: hybrid
  densely connected unet for liver and tumor segmentation from ct volumes,''
  \emph{IEEE transactions on medical imaging}, vol.~37, no.~12, pp. 2663--2674,
  2018.

\bibitem{dong2020multi}
H.~Dong, J.~Pan, L.~Xiang, Z.~Hu, X.~Zhang, F.~Wang, and M.-H. Yang,
  ``Multi-scale boosted dehazing network with dense feature fusion,'' in
  \emph{Proceedings of the IEEE/CVF Conference on Computer Vision and Pattern
  Recognition}, 2020, pp. 2157--2167.

\bibitem{isola2017image}
P.~Isola, J.-Y. Zhu, T.~Zhou, and A.~A. Efros, ``Image-to-image translation
  with conditional adversarial networks,'' in \emph{Proceedings of the IEEE
  conference on computer vision and pattern recognition}, 2017, pp. 1125--1134.

\bibitem{2020SciPy-NMeth}
P.~{Virtanen}, R.~{Gommers}, T.~E. {Oliphant}, M.~{Haberland}, T.~{Reddy},
  D.~{Cournapeau}, E.~{Burovski}, P.~{Peterson}, W.~{Weckesser}, J.~{Bright},
  S.~J. {van der Walt}, M.~{Brett}, J.~{Wilson}, K.~{Jarrod Millman},
  N.~{Mayorov}, A.~R.~J. {Nelson}, E.~{Jones}, R.~{Kern}, E.~{Larson},
  C.~{Carey}, {\.I}.~{Polat}, Y.~{Feng}, E.~W. {Moore}, J.~{Vand erPlas},
  D.~{Laxalde}, J.~{Perktold}, R.~{Cimrman}, I.~{Henriksen}, E.~A. {Quintero},
  C.~R. {Harris}, A.~M. {Archibald}, A.~H. {Ribeiro}, F.~{Pedregosa}, P.~{van
  Mulbregt}, and S.~.~. {Contributors}, ``{SciPy 1.0: Fundamental Algorithms
  for Scientific Computing in Python},'' \emph{Nature Methods}, vol.~17, pp.
  261--272, 2020.

\bibitem{gvtulder2019}
G.~van Tulder, ``elasticdeform,''
  \url{https://github.com/gvtulder/elasticdeform}, 2019.

\bibitem{wang2004image}
Z.~Wang, A.~C. Bovik, H.~R. Sheikh, and E.~P. Simoncelli, ``Image quality
  assessment: from error visibility to structural similarity,'' \emph{IEEE
  transactions on image processing}, vol.~13, no.~4, pp. 600--612, 2004.

\bibitem{hore2010image}
A.~Hore and D.~Ziou, ``Image quality metrics: Psnr vs. ssim,'' in \emph{2010
  20th International Conference on Pattern Recognition}.\hskip 1em plus 0.5em
  minus 0.4em\relax IEEE, 2010, pp. 2366--2369.

\bibitem{zhang2018unreasonable}
R.~Zhang, P.~Isola, A.~A. Efros, E.~Shechtman, and O.~Wang, ``The unreasonable
  effectiveness of deep features as a perceptual metric,'' in \emph{Proceedings
  of the IEEE conference on computer vision and pattern recognition}, 2018, pp.
  586--595.

\bibitem{krizhevsky2012imagenet}
A.~Krizhevsky, I.~Sutskever, and G.~E. Hinton, ``Imagenet classification with
  deep convolutional neural networks,'' in \emph{Advances in neural information
  processing systems}, 2012, pp. 1097--1105.

\bibitem{he2016deep}
K.~He, X.~Zhang, S.~Ren, and J.~Sun, ``Deep residual learning for image
  recognition,'' in \emph{Proceedings of the IEEE conference on computer vision
  and pattern recognition}, 2016, pp. 770--778.

\bibitem{huang2017densely}
G.~Huang, Z.~Liu, L.~Van Der~Maaten, and K.~Q. Weinberger, ``Densely connected
  convolutional networks,'' in \emph{Proceedings of the IEEE conference on
  computer vision and pattern recognition}, 2017, pp. 4700--4708.

\bibitem{zhang2018chromosome}
W.~Zhang, S.~Song, T.~Bai, Y.~Zhao, F.~Ma, J.~Su, and L.~Yu, ``Chromosome
  classification with convolutional neural network based deep learning,'' in
  \emph{2018 11th International Congress on Image and Signal Processing,
  BioMedical Engineering and Informatics (CISP-BMEI)}.\hskip 1em plus 0.5em
  minus 0.4em\relax IEEE, 2018, pp. 1--5.

\bibitem{qin2019varifocal}
Y.~Qin, J.~Wen, H.~Zheng, X.~Huang, J.~Yang, N.~Song, Y.-M. Zhu, L.~Wu, and
  G.-Z. Yang, ``Varifocal-net: A chromosome classification approach using deep
  convolutional networks,'' \emph{IEEE transactions on medical imaging},
  vol.~38, no.~11, pp. 2569--2581, 2019.

\bibitem{zhang2018self}
H.~Zhang, I.~Goodfellow, D.~Metaxas, and A.~Odena, ``Self-attention generative
  adversarial networks,'' \emph{arXiv preprint arXiv:1805.08318}, 2018.

\bibitem{wang2018high}
T.-C. Wang, M.-Y. Liu, J.-Y. Zhu, A.~Tao, J.~Kautz, and B.~Catanzaro,
  ``High-resolution image synthesis and semantic manipulation with conditional
  gans,'' in \emph{Proceedings of the IEEE conference on computer vision and
  pattern recognition}, 2018, pp. 8798--8807.

\end{thebibliography}

\end{document}